\documentclass[fleqn,10pt]{wlscirep}
\usepackage[utf8]{inputenc}
\usepackage[T1]{fontenc}
\usepackage{lineno}
\usepackage{cleveref}

\usepackage{acronym}
\usepackage{siunitx}
\newacro{ctc}[CTC]{Connectionist Temporal Classifier}
\newacro{htr}[HTR]{Handwritten Text Recognition}
\newacro{cer}[CER]{Character Error Rate}
\newacro{wer}[WER]{Word Error Rate}
\newacro{cnn}[CNN]{Convolutional Neural Network}
\newacro{lstm}[LSTM]{Long Short-Term Memory}
\newacro{aer}[AER]{Abbreviation Error Rate}
\newacro{pe}[PE]{positional encoding}
\newacro{crnn}[CRNN]{Convolutional Recurrent Neural Network}
\newacro{wi}[WI]{Writer Identification}

\newcommand{\longs}{\raisebox{-0.55ex}{\includegraphics[height=1em]{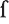}}}
\newcommand{\zcauda}{\raisebox{-0.55ex}{\includegraphics[height=1em]{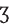}}}

\title{Nuremberg Letterbooks: A Multi-Transcriptional Dataset of Early 15th Century Manuscripts for Document Analysis}

\author[1,*]{Martin Mayr}
\author[2]{Julian Krenz}
\author[3]{Katharina Neumeier}
\author[4]{Anna Bub}
\author[4]{Simon Bürcky}
\author[1,2,3,4]{Nina Brolich}
\author[2]{Klaus Herbers}
\author[3]{Mechthild Habermann}
\author[4]{Peter Fleischmann}
\author[1]{Andreas Maier}
\author[1]{Vincent Christlein}
\affil[1]{Pattern Recognition Lab, Friedrich-Alexander-Universität Erlangen-Nürnberg, Erlangen, 91058, Germany}
\affil[2]{Senior Fellow of Medieval History, Friedrich-Alexander-Universität Erlangen-Nürnberg, Erlangen, 91054, Germany}
\affil[3]{Department of German Linguistics, Friedrich-Alexander-Universität Erlangen-Nürnberg, Erlangen, 91054, Germany}
\affil[4]{Chair of Regional History of Bavaria and Franconia, Friedrich-Alexander-Universität Erlangen-Nürnberg, Erlangen, 91054, Germany}

\affil[*]{corresponding author: Martin Mayr (martin.mayr@fau.de)}

\begin{abstract}
Most datasets in the field of document analysis utilize highly standardized labels, which, while simplifying specific tasks, often produce outputs that are not directly applicable to humanities research. 
In contrast, the Nuremberg Letterbooks dataset, which comprises historical documents from the early 15th century, addresses this gap by providing multiple types of transcriptions and accompanying metadata.
This approach allows for developing methods that are more closely aligned with the needs of the humanities.
The dataset includes 4 books containing 1711 labeled pages written by 10 scribes. 
Three types of transcriptions are provided for handwritten text recognition: Basic, diplomatic, and regularized. For the latter two, versions with and without expanded abbreviations are also available.
A combination of letter ID and writer ID supports writer identification due to changing writers within pages.
In the technical validation, we established baselines for various tasks, demonstrating data consistency and providing benchmarks for future research to build upon.
\end{abstract}
\begin{document}

\flushbottom
\maketitle

\thispagestyle{empty}

\begin{figure}
    \centering
    \includegraphics[width=1.\linewidth]{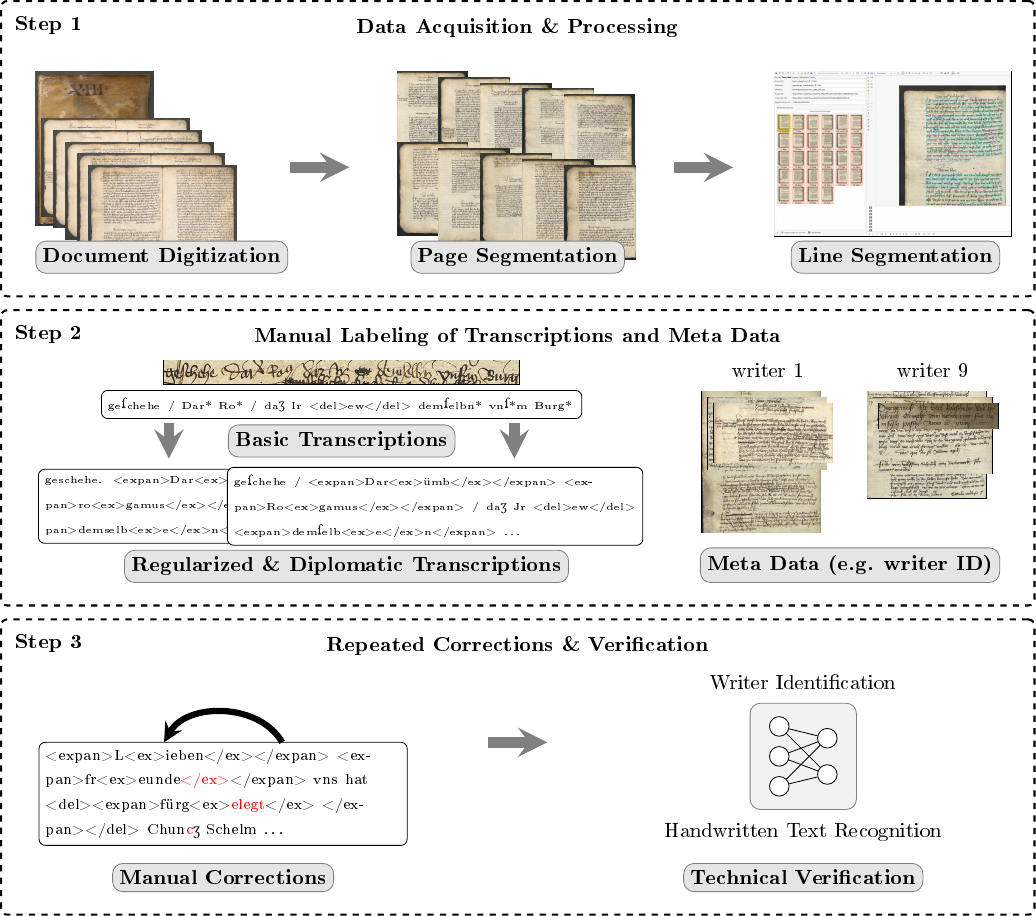}
    \caption{Step 1 describes data acquisition and preprocessing. The pages of the scanned documents are separated, with a subsequent line segmentation. In step 2, the transcriptions and meta data are manually labeled. The created basic transcriptions are used as a foundation for the regularized and diplomatic text versions. Simultaneously, the meta data, like writer IDs, are marked. In step 3, manual corrections are made, and the produced data is analyzed for technical validation.}
    \label{fig:overview}
\end{figure}

\section*{Background \& Summary}

In historical document digitization and handwritten text recognition, a significant challenge lies in bridging the gap between merely scanning ancient manuscripts and truly accessing and understanding the corpus they present. While digitization has made these texts more available, it does not inherently make them comprehensible or usable for varied research purposes. The process of transitioning from a physical document to a digital corpus encompasses numerous complexities, especially in transcription and interpretation, which vary widely based on the research field and study objectives.

\begin{itemize}
\item Computer scientists, for instance, often work with basic transcriptions of documents. These transcriptions are simplified versions tailored for training text recognition models.
\item In contrast, German studies require transcriptions that are as close as possible to the original text, capturing its visual features and nuances. This level of detail is crucial for studies focused on linguistic and cultural contexts. 
\item Historians, on the other hand, lean towards regularized versions of texts where abbreviations are resolved and special characters like ``\zcauda{}'' are normalized to suit the reading habits of the contemporary audience. This approach facilitates content analysis and interpretation, making historical texts more accessible and understandable.  These regularized versions often culminate in the creation of scholarly editions.
\end{itemize}

The Nuremberg Letterbooks illustrate the critical need for varied transcription methods in document analysis. Historically, they were used to record the outgoing letters of Nuremberg's small council to other cities and individuals. The topics of the correspondence range from everyday economic or legal matters of individual citizens to discussions of imperial politics with the kings or other major cities.
An interdisciplinary project team worked collaboratively on some of the oldest preserved books. In this project, three types of annotations have been made available for four successive books, each tailored to meet the specific needs of different research fields. 
The basic transcription is primarily intended for automatic text recognition and often serves as the default in other datasets.\cite{Toseli15Bentham,Sanchez15Bentham,Sanchez16Bozen,Sanchez17Escher} %
The diplomatic and regularized transcriptions cater to the needs of scholars in German Studies and historians, respectively. Additionally, the dataset includes information on expanded abbreviations, as well as metadata such as writer ID and the starting and ending points of letters.

With its diverse annotations, this dataset is not just a valuable resource for historical research but also serves as a starting point for developing models that can handle different types of transcriptions. As the volume of scanned documents from archives continues to grow, the demand for diverse automatic transcription methods becomes increasingly important. 
Such models can greatly assist humanities researchers, providing them with access to digitized texts without requiring additional adjustments to fit each field's specific requirements. 
Ultimately, this dataset seeks to bridge the gap between the vast quantity of digitized historical texts and the complex, varied demands of humanities research, thereby enabling a more profound and streamlined exploration of our history.

\section*{Methods}

\Cref{fig:overview} outlines our study's methodology, broken down into (1) data acquisition and processing, (2) manual labeling of transcriptions and meta data, and (3) repeated corrections and technical verification.

\subsection*{Data Acquistion and Processing}

The process began by scanning the relevant double pages from the historical documents. 
The pages were segmented semi-automatically to isolate each page from the double-page scan.

To achieve accurate segmentation, we employed a Sobel kernel \cite{Kanopoulos:Sobel:1988} in the vertical direction to compute gradients, highlighting the edges within the image. 
These gradients were then summed along the vertical direction and multiplied by a Gaussian curve across the width. 
The focus on the center regions of the double scan is necessary due to false classifications of the borders of the text blocks as page breaks and the prior knowledge that each page's boundary is generally located near the center of the scan. 
This initial segmentation process was further refined by applying small rotations to the image, ultimately selecting the highest gradient activation orientation. 
Each page was manually checked after the automated segmentation process because an error in this early stage would have propagated throughout all successive stages.
After obtaining single pages, we utilized the line segmentation feature CITLab advanced~\cite{Gruening2019Line} of Transkribus (\url{https://www.transkribus.org/}), a specialist text recognition and transcription tool. 
It provided a solid foundation for the baselines and polygons of the text lines. However, strike-throughs and other uncommon text formatting often led to misclassifications, which were adjusted in the next step.

\subsection*{Manual Labeling of Transcriptions and Meta Data}

Once the lines were segmented, experts in German studies and historians, who specialized in documents from this era, created the basic transcriptions.
Simultaneously, rough adjustments to the bounding boxes were made if necessary.
Subsequently, we extended the basic transcriptions to create a diplomatic version through a semi-automatic process that expanded the abbreviations and incorporated known elements like specific symbols. 
This diplomatic version closely reflects the original text, preserving its unique features and stylistic elements.

In addition to the diplomatic version, we created a regularized version of the text derived from the basic and diplomatic transcriptions through a semi-automatic process.
This step included modifications such as converting the long "s" (\longs{}) to the modern "s", enhancing the text's readability for contemporary audiences. 

\subsection*{Repeated Corrections and Technical Verification}

The methodology was completed with multiple rounds of corrections and technical verification to ensure the accuracy and consistency of the transcriptions and metadata across the dataset. 
The technical validation included measuring the performance with handwritten text recognition and writer identification systems. 
Each verification round involved cross-checking the transcriptions and metadata with the original documents and making necessary adjustments.

\begin{table}[ht]
\centering
\begin{tabular}{|l|r|r|r|r|}
\hline
 & \# of pages & \# of lines (basic) & \# of lines (diplomatic) \\
\hline
book 2 (February 7, 1408 – August 12, 1409) & 256  & 8426 & 7997 \\
\hline
book 3 (August 20, 1409 – November 5, 1412) & 548  & 16183 & 15471 \\
\hline
book 4 (June 6, 1414 – April 18, 1416) & 290 & 8439 & 8134 \\
\hline
book 5 (April 21, 1419 – March 31, 1423) & 617 & 17932 & 16720 \\
\hline
\end{tabular}
\caption{\label{tab:book_overview}Note, the counting of the books is based on the preserved ones (6 books between around 1385 and March 1404 are lost), and book~1 (1404 March 21 – 1408 February 8) was subject to another project. Note that there is a slight difference between the number of basic lines and diplomatic lines due to transcription rules, where interlinears are incorporated into the main line.}
\end{table}

\section*{Data Records}

The Nuremberg Letterbooks dataset (\url{https://zenodo.org/records/13881575}) is publicly available and hosted on the CERN-supported open repository Zenodo.\cite{Mayr:zenodo:2024}
\Cref{tab:book_overview} provides an overview of the different books. 
Basic transcriptions are annotated using the PAGE XML format, with each page in the dataset corresponding to a PAGE XML file that includes information such as line polygons (which define the text regions), writer IDs (which identify the main writers of the letter), and basic transcription. Additional PAGE XML files provide both diplomatic and regularized transcriptions. The diplomatic version is linked to line polygons and writer IDs, while the regularized version is not linked to individual lines but instead presents normalized text at the letter level.
Overall, the dataset consists of 17932 annotated lines with basic transcriptions and 16720 lines with diplomatic transcriptions. 
Additionally, the main writers of each letter were tracked for all the books. 
In books 2 through 5, ten main writers created the manuscripts. \Cref{tab:writer_distribution} shows the total number of main writers per book. In book 2, the main writer was Writer 1; in book 3, it was Writer 7; and in books 4 and 5, Writer 9 became the main writer. 
It should be noted that some writer IDs (W3-W5) are absent from this dataset, as they only appeared in book 1.
To facilitate access to the data, we have made our data loading and models available in separate GitHub repositories.

\begin{table}[ht]
\centering
\begin{tabular}{|l|r|r|r|r|r|r|r|r|r|r|r|}
\hline
 & \multicolumn{10}{c}{Writers} & \\
\cline{2-12}
 & W1 & W2 & W6 & W7 & W8 & W9 & W10 & W11 & W12 & W13 & total \\
\hline
book 2 & 247 & 46 & 52 & 125 & 1 & - & - & - & - & - & 471 \\
\hline
book 3 & 348 & 75 & 76 & 494 & 74 & - & - & - & - & - & 1067 \\
\hline
book 4 & 138 & 8 & - & - & 36 & 375 & - & - & - & 1 & 558 \\
\hline
book 5 & 302 & 2 & - & - & - & 764 & 13 & 6 & 4 & - & 1091 \\
\hline
total & 1035 & 131 & 128 & 619 & 111 & 1139 & 13 & 6 & 4 & 1 & 3187 \\
\hline
\end{tabular}
\caption{\label{tab:writer_distribution}Writer distribution across the different books. Note that some writers are not present in some books.}
\end{table}

\section*{Technical Validation}

We applied handwritten text recognition approaches to validate the basic and diplomatic transcriptions. 
Additionally, we used an unsupervised approach from writer identification to check the writer IDs. 

\subsection*{Basic and Diplomatic Transcriptions}

\begin{figure}
    \centering
    \includegraphics[width=0.8\linewidth]{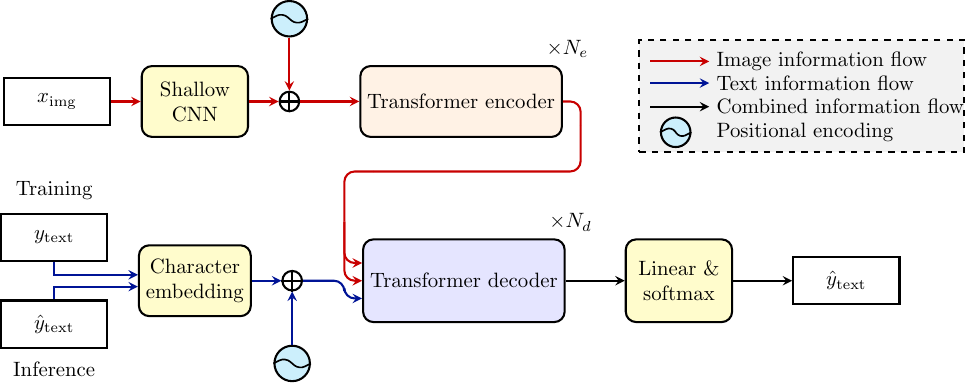}
    \caption{Overview of the Handwritten Text Recognition model. Red arrows show the image information flow, blue arrows show the text information flow, and black arrows show the combined information flow. The architecture is a combination of a shallow CNN and a transformer.}
    \label{fig:htr_architecture}
\end{figure}

We employed a well-established sequence-to-sequence architecture to measure overall performance and manually check the errors for the basic and diplomatic transcriptions.
The architecture of the model is depicted in \cref{fig:htr_architecture}, consisting of a shallow CNN~\cite{wick2021bidirectional, wick2022rescoring} and a transformer network.~\cite{vaswani2017transformer} 
The CNN extracts the visual features from the image and feeds them as an input sequence to the transformer encoder. Based on the previous tokens and the transformer outputs, the decoder predicts the output sequence auto-regressively. For fast training, teacher forcing was applied.

\acf{cer} and \ac{wer} are commonly used for measuring the performance of \ac{htr} tasks.\cite{chaudhary2022easter2,kang2022pay} It is important to note that words are separated solely by the \textit{space} symbol; punctuation marks are included in the metric and may be present within words.
Additionally, we assessed the accuracy of the automatically expanded abbreviations by introducing \acf{aer}, which is defined as:

\begin{equation}
\text{\acs{aer}} = \frac{\text{\# of false abbreviations}}{\text{\# of all abbreviations}}\;.
\end{equation}

We chose to measure if the complete expanded abbreviation is correct, including the position of the surrounding tags.

\begin{table}[]
\centering
\begin{tabular}{|l|l|l|r|r|r|r|}
\hline
Train sets & Validation set & Test set & $\text{CER}_{\text{val}}$ & $\text{WER}_{\text{val}}$ & $\text{CER}_{\text{test}}$ & $\text{WER}_{\text{test}}$ \\
\hline
book 3, book 5 & book 4 & book 2 & $2.64$ & $9.80$ & $3.49$ & $11.65$ \\
\hline
book 2, book 5 & book 4 & book 3 & $2.98$ & $10.91$ & $3.71$ & $12.25$ \\
\hline
book 3, book 5 & book 2 & book 4 & $3.49$ & $11.64$ & $2.88$ & $10.61$ \\
\hline
book 2, book 3 & book 4 & book 5 & $5.88$ & $20.73$ & $7.06$ & $24.09$ \\
\hline
\end{tabular}
\caption{\label{tab:htr_basic_results} \ac{htr} results on basic transcriptions. Results are given in percent~[\%].}
\end{table}

\begin{table}[]
\centering
\begin{tabular}{|l|l|l|r|r|r|r|}
\hline
Train sets & Validation set & Test set & $\text{CER}_{\text{val}}$ & $\text{WER}_{\text{val}}$ & $\text{CER}_{\text{test}}$ & $\text{WER}_{\text{test}}$ \\
\hline
book 3, book 5 & book 4 & book 2 & $2.27$ & $8.49$ & $4.14$ & $14.74$ \\
\hline
book 2, book 5 & book 4 & book 3 & $2.47$ & $9.13$ & $3.94$ & $13.47$ \\
\hline
book 3, book 5 & book 2 & book 4 & $4.06$ & $14.42$ & $2.41$ & $9.01$ \\
\hline
book 2, book 3 & book 4 & book 5 & $6.61$ & $23.56$ & $7.11$ & $25.40$ \\
\hline
\end{tabular}
\caption{\label{tab:htr_dipl_results} \ac{htr} results on diplomatic transcriptions. Results are given in percent~[\%].}
\end{table}

\begin{table}[]
\centering
\begin{tabular}{|l|l|l|r|r|r|r|r|r|}
\hline
Train sets & Validation set & Test set & $\text{CER}_{\text{val}}$ & $\text{WER}_{\text{val}}$ & $\text{AER}_{\text{val}}$ & $\text{CER}_{\text{test}}$ & $\text{WER}_{\text{test}}$ & $\text{AER}_{\text{test}}$ \\
\hline
book 3, book 5 & book 4 & book 2 &  $2.65$ & $9.46$ & $5.87$ & $4.75$ & $15.39$ & $11.29$ \\
\hline
book 2, book 5 & book 4 & book 3 & $2.83$ & $10.17$ & $6.68$ & $4.56$ & $14.17$ & $10.95$ \\
\hline
book 3, book 5 & book 2 & book 4 & $4.86$ & $15.38$ & $11.22$ & $2.82$ & $9.94$ & $6.40$ \\
\hline
book 3, book 2 & book 4 & book 5 & $7.72$ & $25.64$ & $17.84$ & $8.64$ & $28.10$ & $20.44$ \\
\hline
\end{tabular}
\caption{\label{tab:htr_dipl_abbr_results} \ac{htr} results on diplomatic transcriptions with expanded abbreviations. Results are given in percent~[\%].}
\end{table}

\Cref{tab:htr_basic_results} shows that all the runs perform reasonably well. When using book 4 for testing, the lowest \ac{cer} and \ac{wer} were achieved. book 2 and book 3, as test sets, performed similarly. However, book 5 had the highest \ac{cer} and \ac{wer} with $7.06\%$ and $24.09\%$, respectively. Due to also worse validation results, we hypothesize that the diverse writer distribution of book 5 is crucial to achieving very good recognition results on the other books. 

A similar trend can be seen in \cref{tab:htr_dipl_results}, which shows the results of the \ac{htr} model on the diplomatic transcriptions. As expected, due to the more complex data, on average the \ac{cer} and \ac{wer} are slightly increased compared to the basic transcriptions. Again, book 4 was the easiest to predict, and book 5 was the hardest. The \ac{cer} and \ac{wer} computed from the results of book 4 were even lower than those of the basic results.

\Cref{tab:htr_dipl_abbr_results} depicts the recognition results with extended abbreviations. \ac{cer} and \ac{wer} dropped slightly because of longer output sequences, but they were still in a similar error range. Interestingly, the completion of the abbreviations worked very well. Note that every part of the expanded abbreviation must be correctly predicted for a correct recognition.
We hypothesize that the transformer decoder's implicit language model receives enough samples from the training data to replace the abbreviation symbol with the appropriate text automatically.

\subsection*{Writer Information}

Following the approach outlined by Christlein et al.,~\cite{Christlein15ICDAR,Christlein17PR,Christlein18DAS} we employed a widely recognized writer retrieval pipeline for the \ac{wi} task. 
First, the input images are binarized with Otsu thresholding. \cite{otsu1979thresholding} 
For retrieving local feature vectors, RootSIFT descriptors~\cite{Arandjelovic12} were computed from SIFT keypoints,~\cite{Lowe04} which were then jointly whitened and reduced in dimensionality using PCA.~\cite{Christlein17PR}
The global feature vector is the result of a multi-VLAD approach,\cite{Arandjelovic2013mvlad} where multiple VLAD encodings are concatenated.
For a better generalization, the global feature representation is PCA whitened.~\cite{Christlein15ICDAR}

The score was determined using leave-one-sample-out cross-validation, where each sample is picked as a query, and the remaining samples are ranked according to their similarity to the query. Mean Average Precision (mAP) is computed from the ranks. We also report the top-1 accuracy.
We give the results for each book. Despite the approach's unsupervised nature, we split the test data from the train data.

\begin{table}[]
\centering
\begin{tabular}{|l|l|r|r|}
\hline
Train sets & Test set & Top-1 & mAP \\
\hline
book 3, book 4, book 5 & book 2 &  $0.996$ & $0.928$ \\
\hline
book 2, book 4, book 5 & book 3 & $0.993$ & $0.883$ \\
\hline
book 2, book 3, book 5 & book 4 & $0.995$ & $0.938$ \\
\hline
book 2, book 3, book 4 & book 5 & $0.996$ & $0.964$ \\
\hline
All books & All books & $0.993$ & $0.857$ \\
\hline
\end{tabular}
\caption{\label{tab:writer_results} Writer identification results.}
\end{table}

\begin{figure}
    \centering
    \includegraphics[width=1.\linewidth]{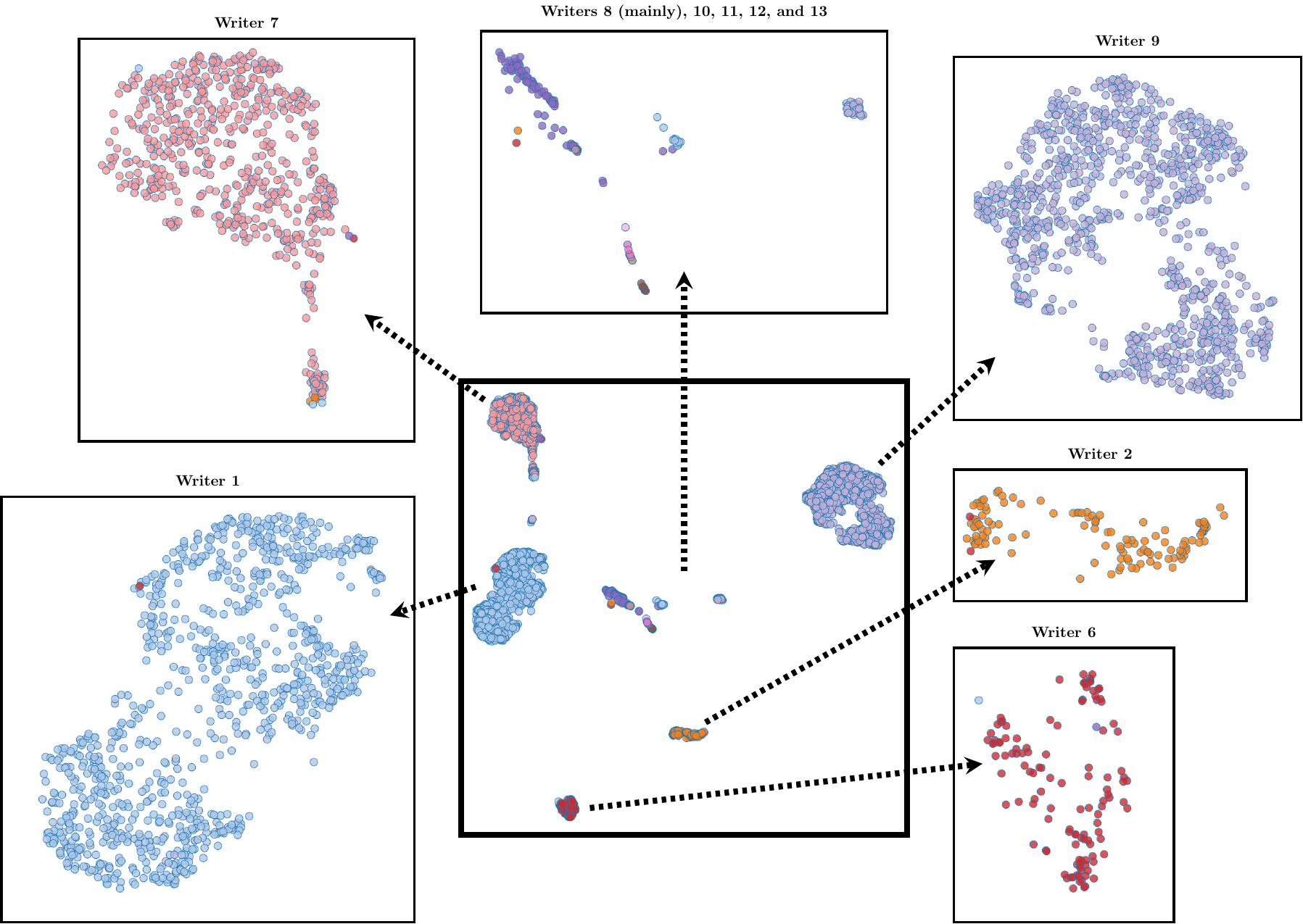}
    \caption{Visualization of dimensionality reduced global feature vectors of all books. Each sample point denotes one letter in the letterbooks and is color-coded by the specific writer label. The box in the middle gives an overview of all samples. The outgoing boxes are zoomed-in versions of writer clusters.}
    \label{fig:wi_umap_results}
\end{figure}

\Cref{tab:writer_results} gives the results for a book-wise 4-fold-cross validation. The top-1 scores were almost perfect. However, when using book 3 for testing, mAP dropped below $90\,\%$. The best mAP was achieved for book 5 with $96.4\,\%$
Also, for using all books as a basis for applying the unsupervised training method and also for testing, the results were still very good despite the more extensive test set with a top-1 score and mAP of $99.3\,\%$ and $85.7\,\%$, respectively.  

\Cref{fig:wi_umap_results} visualizes the global feature vectors with a UMap dimensionality reduction.~\cite{mcinnes2020umap} In the middle, the total view of the outputs is given. Each sample point represents a letter of the dataset written by a specific writer that is color-coded. 
Around the middle figure, the most prominent clusters are zoomed in to show how clean the predictions are.
The top box is used for the underrepresented clusters. The intra-class distance was still as expected, but the inter-class distance was much lower than for the most frequent writers.
Upon closer inspection of the outliers, the experts recognized that it was not a single writer but often multiple writers who created the misclassified letters. Overall, the labels for the main writers are still correct; however, the issue lies in the automatic approach, which in these cases focuses on the wrong parts of the letter.

\subsection*{Summary}
Overall, the technical validation demonstrates strong and consistent results for both handwritten text recognition and writer identification tasks, underscoring the robustness of our dataset and validation methodologies.
Including various transcription types (basic, diplomatic, and regularized) supports a more nuanced analysis of historical documents, which is particularly valuable for humanities research that requires high fidelity to original sources.
With the additional manual checks of error cases, the dataset is well-suited to serve as a foundation for developing approaches that align more closely with the specific needs of humanities scholars. 

Future work could leverage models trained on this dataset to semi-automatically process further books in the Nuremberg Letterbooks series, enhancing the accessibility of this rich historical content for humanities researchers and enabling more in-depth exploration. Moreover, given the dataset's size and variability, document analysis experts could investigate the application of Large Language Models and Vision-Language Models to further advance the analysis of this type of historical data.

\newpage
\section*{Code availability}\label{sec:code}

Code of the semi-automatic data preparation and for loading the data and rerunning the experiments is publicly available:
\begin{itemize}
    \item Data Preparation: \url{https://github.com/M4rt1nM4yr/letterbooks_data_preparation}
    \item Handwritten Text Recognition: \url{https://github.com/M4rt1nM4yr/letterbooks_text_verification}
    \item Writer Identification: \url{https://github.com/M4rt1nM4yr/letterbooks_writer_verification}
\end{itemize}

\bibliography{sample}

\section*{Acknowledgements}

\begin{itemize}
	\item We acknowledge funding by the Deutsche Forschungsgemeinschaft (DFG, German Research Foundation) -- 416910787.  
	\item The authors gratefully acknowledge the scientific support and HPC resources provided by the Erlangen National High Performance Computing Center (NHR@FAU) of the Friedrich-Alexander-Universität Erlangen-Nürnberg (FAU). The hardware is funded by the German Research Foundation (DFG).
\end{itemize}

\section*{Author contributions statement}
All authors contributed to the study's conception and design. M.M. conducted the experiments and analyzed the results. M.M. wrote the first draft of the manuscript. All authors reviewed and approved the final manuscript.

\section*{Competing interests}

The authors declare no competing interests.

\end{document}